\documentclass[10pt,twocolumn,letterpaper]{article}

\usepackage{cvpr}
\usepackage{times}
\usepackage{epsfig}
\usepackage{amsfonts}
\usepackage{bbding}
\usepackage{graphicx}
\usepackage{amsmath}
\usepackage{amssymb}
\usepackage{algorithm}
\usepackage{multirow}
\usepackage{siunitx}
\usepackage{array}
\usepackage{booktabs}
\usepackage{tabularray}
\usepackage{ulem}
\usepackage{subcaption}
\usepackage{makecell}
\usepackage{algpseudocode}
\usepackage{tabularx}
\usepackage{bm}
\begin{document}
\title{IdentiFace: Multi-Modal Iterative Diffusion Framework for Identifiable Suspect Face Generation in Crime Investigations}

\author{Weichen Liu\\
Southeast University\\
Nanjing, China\\
{\tt\small 213233033@seu.edu.cn}
\and
Yixin Yang\\
Shenzhen MSU-BIT University\\
Shenzhen, China\\
{\tt\small yyx@smbu.edu.cn}
\and
Changsheng Chen\thanks{Corresponding author}\\
Shenzhen MSU-BIT University\\
Shenzhen, China\\
{\tt\small cschen@smbu.edu.cn}
\and
Alex Kot\\
Shenzhen MSU-BIT University\\
Shenzhen, China\\
{\tt\small kcc@smbu.edu.cn}
}

\maketitle
\thispagestyle{empty}

\begin{abstract}
Suspect face generation remains a technical challenge in crime investigations. Traditional sketch-drawing workflows suffer from low efficiency and quality, while diffusion-based approaches still face intrinsic limitations on conditional ambiguity for text-to-image models and sampling variance for one-shot generation. We proposed IdentiFace, a novel diffusion-based framework for identifiable suspect face generation, which addressed these issues through (1) multi-modal input design to strengthen conditional control, and (2) an iterative generation pipeline enabling identifiable feature adjustment. We additionally contributed a facial identity loss and two task-specific datasets. Comprehensive experiments on synthetic datasets and in real-world scenarios indicate that IdentiFace achieves superior performance over existing methods, especially in terms of identity retrieval, and shows strong potential for practical applications. 
\end{abstract}

\section{Introduction}\label{sec:intro}
\begin{figure}[tbp]
\begin{subfigure}{\linewidth}
  \centering
  \includegraphics[width=\linewidth]{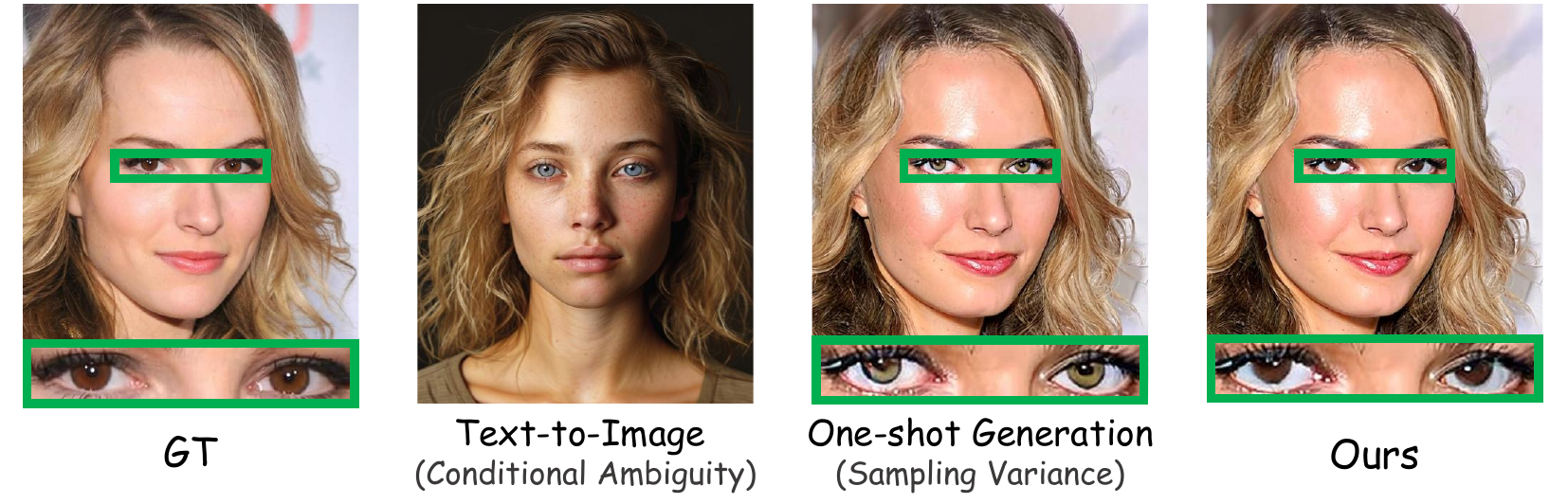}
  \caption{Comparisons between GT, text-to-image (T2I) model~\cite{chen2023pixartalpha}, one-shot generation, and our result. Existing works face limitations on conditional ambiguity (structural guidance shortage for T2I models) and sampling variance (identifiable feature mismatch for one-shot generation), while ours addressed these problems.}
  \label{subfig:introduction1}
\end{subfigure}
\newline
\begin{subfigure}{\linewidth}
  \centering
  \includegraphics[width=\linewidth]{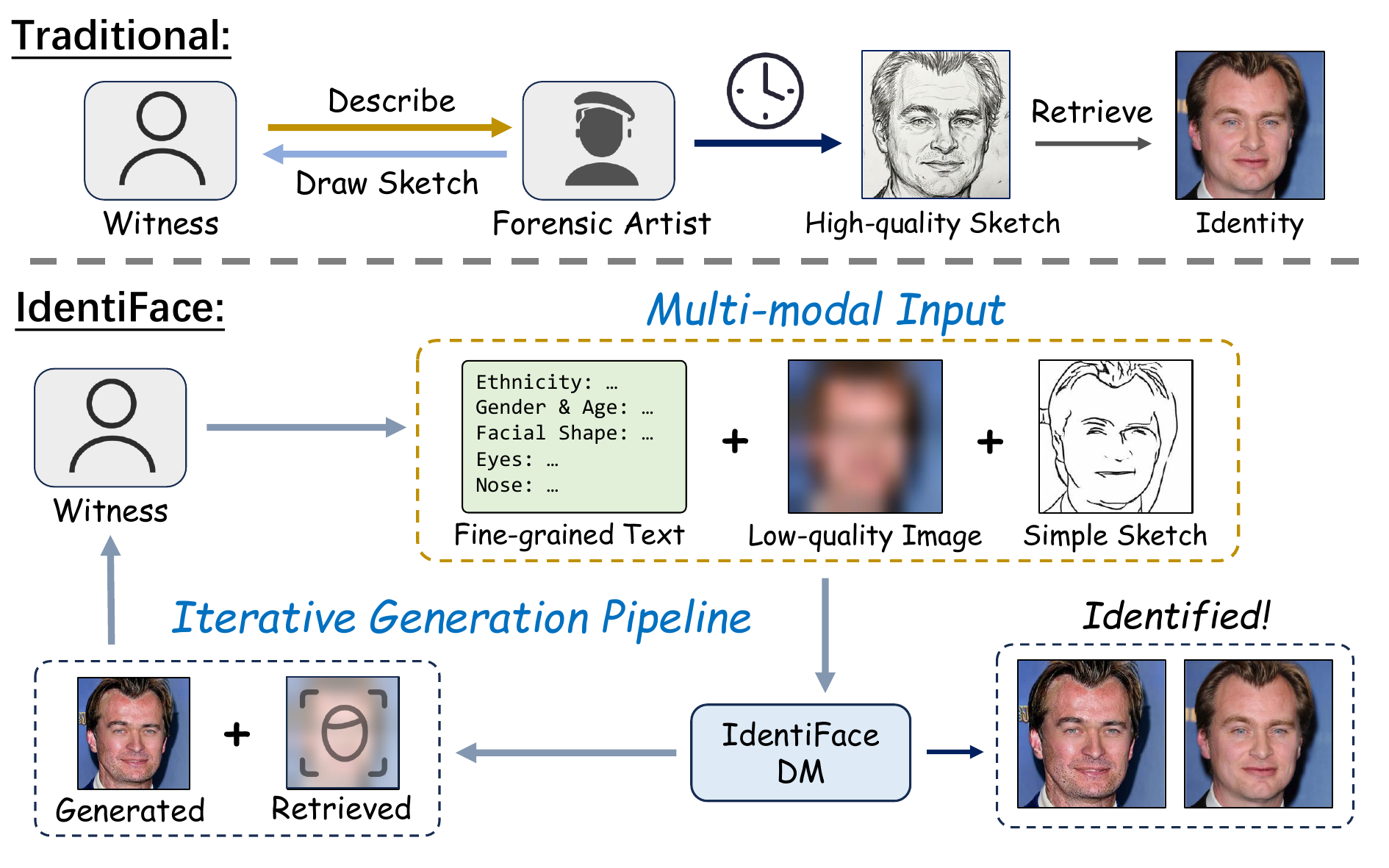}
  \caption{We designed multi-modal input and iterative generation pipeline for identifiable suspect face generation, surpassing traditional workflows in both speed and quality.}
  \label{subfig:introduction2}
\end{subfigure}
\caption{Demonstration of task challenges and our contributions.}
\label{figure_1}
\end{figure}

In crime investigations, reconstructing a suspect’s facial appearance from witness (or victim) memory is a critical yet challenging task. The traditional workflow typically requires a witness providing verbal descriptions of the suspect’s facial features, which are then translated by a trained forensic artist into a composite sketch for suspect identification. Despite its longstanding use in law enforcement, this process is time-consuming, skill-dependent, and produces sketch-like images with limited photorealism.

Recent advances in diffusion models (DMs)~\cite{cai2025z, chen2024pixartdelta, chen2023pixartalpha, ho2020denoising, labs2025flux1kontextflowmatching, rombach2022high, saharia2022imagen, song2020denoising, xie2025sana} have enabled high-fidelity image synthesis. In addition, the widespread use of facial recognition models~\cite{boutros2022sface, deng2019arcface, kim2022adaface} allows law enforcement agencies to identify suspects based on their facial images. These advances lead us to a novel task: leveraging diffusion models to generate high-fidelity, identifiable suspect faces conditioned on information in crime investigations, namely crime-scene evidence and witness memory, such that the generated faces can be correctly identified by facial recognition models. This approach promises to be faster and yield higher-quality results than traditional sketch drawing.

However, applying existing works directly to the task remains problematic. The root cause is that generated images are stochastic realizations of a learned conditional   distribution~\cite{ho2020denoising, song2020denoising}, whose nature results in two fundamental limitations (Fig. \ref{subfig:introduction1}): 

\begin{itemize}
\item[(1)] \textbf{Conditional Ambiguity.} Existing works~\cite{jalan2020suspect, kulkarni2025criminal-gan, ravi2024face, warrier2024criminal-stable} rely solely on textual descriptions, which lack structural guidance for diffusion models to generate accurate facial features. This leads to high conditional ambiguity, causing the distribution to deviate from the witness's intent and produce dissimilar facial images.

\item[(2)] \textbf{Sampling variance.} One‑shot generation of diffusion models suffers from inherent sampling variance: A single realization of image may diverge from witness memory in identifiable features. Multiple rounds of adjustment~\cite{avrahami2023blended, couairon2022diffedit} are required to reach an correctly identified result, but the witness lacks a clear criterion for editing and identification.
\end{itemize}

In this paper, we proposed our method \textbf{IdentiFace} (Fig. \ref{subfig:introduction2}) for identifiable suspect face generation in crime investigations. To address \textbf{conditional ambiguity}, we introduced three types of information available in crime investigations: low‑quality image, simple sketch, and fine‑grained text, serving as multi‑modal conditional inputs. This design provides stronger spatial and geometric constraints than plain text, thereby controlling the conditional distribution more effectively. To overcome limitations on \textbf{sampling variance}, we designed an iterative generation pipeline which in each round retrieves the best-matching face of the generated image from a facial database. The witness inspects both faces and then decides whether to stop (if identified) or which region needs refinement. Such mechanism mitigates the impact of randomness and offers a clear reference anchor for generation guidance. Moreover, we optimized the training loss with a facial identity loss, and built two task-specific datasets ID-CelebA and ID-FFHQ based on established datasets~\cite{karras2019FFHQ, CelebAMask-HQ, xia2021MMCelebA}, featuring multi-modal attributes with fine‑grained text descriptions, a key annotation absent in existing facial datasets. 

To evaluate the performance of our method, we conducted comprehensive experiments on our constructed datasets with two IdentiFace model instances built on Stable Diffusion v1.5~\cite{rombach2022high} and PixArt-$\alpha$~\cite{chen2023pixartalpha}. Compared with T2I baseline~\cite{chen2023pixartalpha}, conditional image baseline~\cite{SimoSerraTOG2018Sketch1, SimoSerraSIGGRAPH2016Sketch2}, face restoration~\cite{wang2025osdface}, sketch-to-image~\cite{peng2023difffacesketch}, other multi-modal~\cite{mou2024t2i} methods and ablated variants, IdentiFace consistently achieves superior performance on identity reconstruction metrics and most image quality assessments, attaining the highest identity matching rate of 84\% on ID‑CelebA and 85\% on ID‑FFHQ. Experiments in real‑world scenarios further confirmed its practical applicability.

Our main contributions are summarized as follows:

\begin{itemize}
\item[(1)] We identified three practical input modalities from crime investigations: low‑quality image, simple sketch, and fine‑grained text, to resolve conditional ambiguity and provide stronger control over the conditional distribution.

\item[(2)] We proposed an iterative generation pipeline that retrieves the best-matching face for reference per iteration, counteracting sampling variance and yielding informed guidance toward identifiable faces.

\item[(3)] We constructed two purpose‑built datasets, ID‑CelebA and ID‑FFHQ, synthesizing fine‑grained textual descriptions unavailable in prior datasets~\cite{karras2019FFHQ, CelebAMask-HQ, wang2008cuhkDataset, xia2021MMCelebA}.
\end{itemize}

\begin{figure*}[htbp]
\begin{center}
   \includegraphics[width=\linewidth]{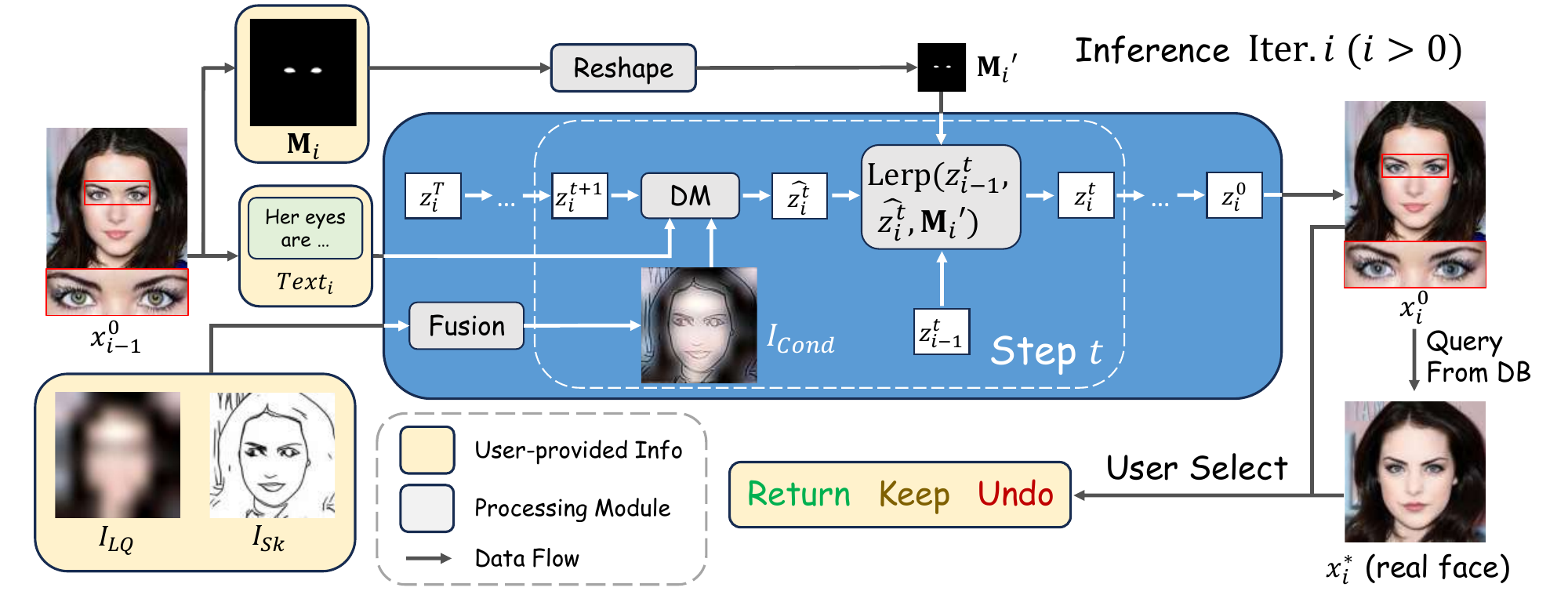}
\end{center}
   \caption{Overview of our proposed method \textbf{IdentiFace}. Our method leverages three modalities of accessible information (low-quality image, sketch image and text) in crime scenes to generate identifiable faces. The two images are fused together as a uni-modal input of ControlNet, providing stronger control for image generation. The iterative generation pipeline enables users to interact with DM round by round. Users provide different masks and texts for DM to edit the facial image in multiple iterations, with the best-retrieval real face presented in each iteration for identification reference. Three choices can be made based on users' perceptions of generation results.} 
\label{fig:overview}
\end{figure*}

\section{Related Works}

\subsection{Diffusion-based Conditional Generation}
Diffusion models (DMs)~\cite{ho2020denoising, song2020denoising} serve as the backbone of numerous T2I image generators~\cite{cai2025z, chen2023pixartalpha, labs2025flux1kontextflowmatching, rombach2022high, saharia2022imagen, xie2025sana}. To incorporate spatial conditions, ControlNet~\cite{zhang2023adding} and T2I-Adapter~\cite{mou2024t2i} enable DMs to process edge maps, depth images, or sketches without retraining the base model; later works~\cite{qin2023unicontrol, zhao2023uni} extend multi-modal visual control. While these methods successfully handle diverse conditional inputs, they target generic image generation and do not address which input modalities available in crime investigations are most suitable for our task. We determined a practical multi‑modal conditioning design to tackle this problem.

\subsection{Image Editing with Localized Control}
For iterative refinement, mask-guided editing methods facilitate precise local re-generation. Blended Diffusion~\cite{avrahami2022blended} provides the first solution for local editing in generic natural images, and Blended Latent Diffusion~\cite{avrahami2023blended} adapts Blended Diffusion to the latent space. Imagen Editor~\cite{wang2023imagenEditor} introduces a cascaded diffusion model fine-tuned from Imagen~\cite{saharia2022imagen} for mask-guided editing, along with a benchmark EditBench. These methods support fine-grained facial feature modifications, but witnesses often cannot provide a clear editing target and direction due to the lack of objective reference. Our pipeline fills this gap by providing the best-matching face, a concrete reference to guide further generation.

\subsection{Facial Generation for Suspect Identification}
Several works have explored generative models for suspect face synthesis. Early attempts~\cite{jalan2020suspect, kulkarni2025criminal-gan, ravi2024face, warrier2024criminal-stable} propose text-to-face generation methods, demonstrating the feasibility of converting witness descriptions into visual outputs. OSDFace~\cite{wang2025osdface} handles low-quality facial images via one-step diffusion restoration. Despite their contributions, none simultaneously addresses two critical challenges of conditional ambiguity and sampling variance, due to their inadequate input modalities or editing instructions. Sketch‑to‑face methods~\cite{que2024denoising, tang2024toward_SKETCH2FACE} achieve identity-preserving generation from high-quality sketches, but obtaining them is time-consuming, limiting practical deployments.

None of the above lines of work provide a concrete solution for identifiable suspect face generation. Our work addresses this by integrating conditional generation and iterative interaction mechanism into a closed‑loop system.

\section{Methods}
\label{sec:methods}
As is discussed in Sec. \ref{sec:intro}, the task of identifiable suspect face generation calls for additional input modalities of accessible information, and an appropriate design of iterative generation mechanism during inference. To address these challenges, we provided \textbf{IdentiFace}, a feasible approach that leverages textual descriptions, simple sketches and extremely low-resolution facial images for generation, and enables mask-guided editing and best-retrieval display to achieve the best identifiable suspect face, with minimal modifications on existing ControlNet-based~\cite{zhang2023adding} diffusion models. We also raised Facial Identity Loss for training of this specialized task. The overview of our proposed method is demonstrated in Fig.\ref{fig:overview}.

\subsection{Multi-modal Conditioning Design}
\label{subsec:modality}
In the task of identifiable suspect face generation, various types of information are available in crime investigations. Relying solely on witness descriptions provides insufficient conditional control, but it has not been discussed what modalities of accessible information can be used for this specific task. 

Images of the suspect's frontal face captured by surveillance cameras are one of the most direct modalities, yet sometimes only extremely low‑quality (LQ) ones are available. Tremendous resolution degradation erases identifiable features, making traditional restoration~\cite{duan2025dit4sr, wang2025osdface} fails and causing the police to discard this clue. We instead introduced LQ image as an information prior, leveraging the general facial shape and color inside.

Since LQ provides limited information, a stronger condition, sketch, is required for additional control. The witness is encouraged by the police to recall the detailed facial features, and draws (or let the forensic artist draw) a simple sketch based on LQ only with a little help. Unlike high-fidelity sketches that can be directly used for identification, the expected ones only include rough positions, sizes and shapes of prominent facial features, e.g., eyes, eyebrows, nose, mouth, ears and hair. Furthermore, the witness gives textual descriptions of the suspect, which guides the perceived gender, age, subtle features, etc., as another conditional input.

Altogether, we proposed three accessible modalities for the generation task: \textbf{LQ} ($I_{LQ}$), \textbf{sketch} ($I_{Sk}$) and \textbf{text} ($Text$), which provide three-level conditional control for generation. The synthesis of these modalities in our dataset is discussed in Sec. \ref{subsec:annotation}.

\begin{figure}[htbp]
\begin{center}
   \includegraphics[width=0.8\linewidth]{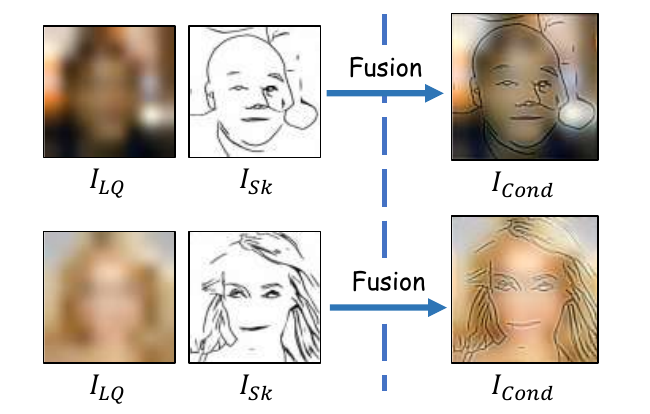}
\end{center}
   \caption{Examples of our proposed LQ, sketch, and fused conditional image.}
\label{fig:fusion}
\end{figure}

For extensibility across different T2I model backbones~\cite{cai2025z, chen2023pixartalpha, labs2025flux1kontextflowmatching, rombach2022high, xie2025sana}, we adopt the backbone-agnostic ControlNet~\cite{zhang2023adding} architecture. Since it expects a single conditional image, we simply fuse $I_{LQ}$ and $I_{Sk}$ into a unified input $I_{Cond}$ (Fig. \ref{fig:fusion}), allowing both modalities to be processed with only one ControlNet branch.

\subsection{Iterative Generation Pipeline}
\label{subsec:iterative_pipeline}
Beyond introducing new modalities for fine-grained control, the interaction between users and generative backbones warrants careful consideration. One-shot face generation remains challenging, as sampling variance~\cite{ho2020denoising, song2020denoising} may lead to misaligned outputs. Moreover, witnesses lack a criterion to align the generated face with their memory to achieve identifiable feature reconstruction.

To address these challenges, we drew inspiration from the traditional workflow between a witness and a forensic artist: The artist produces an initial sketch based on the witness’s first memory, then \textbf{iteratively} improves it based on the witness’s feedback. We transformed such process into a pipeline that gradually refines a facial image based on diffusion models. Algorithm \ref{alg:iterative} describes the iterative generation pipeline of IdentiFace, where $ \text{Lerp}(a,b,c) = b \odot c + a \odot (1-c)$ in Line \ref{algo:lerp}.

\begin{algorithm}[htbp]
\caption{Iterative Facial Image Generation Pipeline}
\label{alg:iterative}
\begin{algorithmic}[1] 
\Require $\text{Fusion}(\cdot)$, $\text{DDIM}_\theta(\cdot)$, $\text{Decode}_\theta(\cdot)$, $\text{DB}$, $F_\Theta(\cdot)$
\Ensure Final facial image $x^0_{\text{final}}$, Top-1 retrieval $x^*_{\text{final}}$

\State $i \gets 0$, $I_{Cond} \gets \text{None}$

\While{True}
    \State $z^{T}_i \gets \text{Noise}(T)$
    \If{$i = 0$}
        \State $(Text_i, I_{LQ}, I_{Sk}) \gets \text{User}()$
        \State $I_{Cond} \gets \text{Fusion}(I_{LQ}, I_{Sk})$
        \For{$t = T-1$ \textbf{downto} $0$}
            \State $z^t_i \gets \text{DDIM}_\theta(z^{t+1}_i, T, t, Text_i, I_{Cond})$
        \EndFor
    \Else
        \State $(\mathbf{M}_i, Text_i) \gets \text{User}()$
        \State $\mathbf{M}'_i \gets \text{Reshape}(\mathbf{M}_i, \text{Shape}(z^T_i))$
        \State Update $I_{Sk}$ and $I_{Cond}$ from User() if needed
        \For{$t = T-1$ \textbf{downto} $0$}
            \State $\widehat{z^t_i} \gets \text{DDIM}_\theta(z^{t+1}_i, T, t, Text_i, I_{Cond})$
            \State \label{algo:lerp} $z^t_i \gets \text{Lerp}(z^t_{i-1}, \widehat{z^t_i}, \mathbf{M}'_i)$
        \EndFor
    \EndIf
    \State $x^0_i \gets \text{Decode}_\theta(z^0_i)$
    \State $x^*_i \gets \arg\max_{x'\in\text{DB}} \cos\bigl[F_\Theta(x'), F_\Theta(x^0_i)\bigr]$
    \State $\text{Display}(x^0_i, x^*_i)$
    \State $c \gets \text{User}() \in \{\text{Return}, \text{Keep}, \text{Undo}\}$

    \If{$c = \text{Return}$}
        \State \Return $(x^0_i, x^*_i)$
    \ElsIf{$c = \text{Keep}$}
        \State $i \gets i+1$
    \EndIf
\EndWhile
\end{algorithmic}
\end{algorithm}
By introducing multiple iterations and best-retrieval demonstrations, our pipeline manages both problems appropriately. Specifically, the pipeline performs mask-guided image editing~\cite{avrahami2023blended, couairon2022diffedit} in iteration $i>0$. Using latent-space mask $\mathbf{M}'_i\in [0,1]^{C'\times H'\times W'}$ reshaped from $\mathbf{M}_i$ explicitly provided by users, the model regenerates only the masked region controlled by $Text_i$, and blends it with the unmasked areas of $z^t_{i-1}$ at every denoising step $t$. The introduction of masks precisely targets the intended region, enabling fine-grained local edits without affecting other features.

After each iteration $i$, the pipeline queries the face database $\text{DB}$ to fetch the best-match facial image
\begin{equation}
\label{eq:top1}
    x^*_i=\arg \max_{x'\in \text{DB}}\!\left(\cos[F_\Theta(x'),F_\Theta(x^0_i)]\right),
\end{equation}
where $F_\Theta(\cdot)$ is a facial feature encoder with a pretrained facial recognition model $\Theta$, and $x_i^0$ is the generated facial image in iteration $i$. Such design enables the user to compare the retrieved face against their memory and decide whether further refinement is needed. 

The user can make three choices based on their perceptions of $x^0_i$ and $x^*_i$:
\begin{itemize}
\item[(1)] \textbf{Return.} Return $x^0_i$ and $x^*_i$ as results, when $x^*_i$ is identified as the suspect or $x^0_i$ is perceived identifiable enough.
\item[(2)] \textbf{Keep.} Keep the results of iteration $i$ and proceed to iteration $i+1$, when $x^0_i$ is acceptable while some features need more editing.
\item[(3)] \textbf{Undo.} Re-execute iteration $i$ using new $Text_i$ and $\mathbf{M}_i$ (if $i>0$), when $x^0_i$ fails to convey the intended features.
\end{itemize}

This generation pipeline itself is training-free, and is easily transferrable to diverse pretrained DM backbones.
\subsection{Facial Identity Loss}
\label{subsec:loss}
The training of IdentiFace mainly follows the traditional ControlNet paradigm. However, the original loss function is designed for general image generation, which is not intrinsically aligned with facial identity reconstruction. To bridge this gap, we developed Facial Identity Loss $\mathcal{L}_{ID}$ with considerations of the application scenarios of this task.

The core idea of $\mathcal{L}_{ID}$ is that the similarity between predicted faces and GT faces should control the gradient flow. To obtain the predicted clean image $z^0_{pred}$, we transformed the forward diffusion formula~\cite{ho2020denoising} into 
\begin{equation}
    z^0_{pred}=\frac{z^t-\sqrt{1-\bar{\alpha}_t}\cdot\epsilon_{\theta}(z^t,t)}{\sqrt{\bar{\alpha}_t}},
\end{equation}
where $\bar{\alpha}_t$ is a hyperparameter defined in~\cite{ho2020denoising}. Because the predicted clean image $z^0_{pred}$ is more stable when derived from late-step latents, we compute $\mathcal{L}_{ID}$ exclusively when the randomly chosen denoising timestep $t<0.2T$.

We then obtained $x^0_{pred}$ by decoding $z^0_{pred}$. The cosine similarity between it and GT face $x^0$ is 
\begin{equation}
\label{eq:similarity}
    \text{Sim}=\cos[F_\Theta(x^0_{pred}),F_\Theta(x^0)], 
\end{equation}
and the Facial Identity Loss is defined as
\begin{equation}
  \mathcal{L}_{ID} = \sigma\big(k(\mathcal{T}_{\Theta} - \text{Sim})\big),
\end{equation}
where $\sigma(\cdot)$ is the Sigmoid function, $k$ is the slope hyperparameter, and $\mathcal{T}_{\Theta}$ is the cosine similarity threshold for model $\Theta$ to determine same-identity faces. 

Instead of merely using $(1-\text{Sim})$ as our loss, we pass $(\mathcal{T}_{\Theta} - \text{Sim})$ through a Sigmoid function, which scales the gradient contribution based on the distance from current similarity to the identity threshold. In other words, $\mathcal{L}_{ID}$ provides strong contributions when the model is still unable to reconstruct identifiable features through conditions, but as the similarity of identity reaches an acceptable threshold, its influence on training loss diminishes quickly.

\section{Dataset Construction}
\label{sec:dataset}
We based our construction on two established high-quality face repositories: CelebAMask-HQ~\cite{karras2017CelebAHQ, CelebAMask-HQ, liu2015CelebA} and FFHQ~\cite{karras2019FFHQ}. CelebAMask-HQ contains 30,000 celebrity facial images with identity annotations, and FFHQ comprises 70,000 high-quality images crawled from Flickr. 

However, neither of them contains fine-grained descriptive text, extremely downgraded LQ and simple sketch image. In addition, the number of images unsuitable for our training and evaluation is non-negligible, and the identity file of FFHQ is absent. To align with our research scenario, we processed these raw datasets to create \textbf{ID-CelebA} and \textbf{ID-FFHQ}.Table \ref{tab:dataset} lists their basic information. Images are saved as .jpg considering disk space, but .png images are still available in original datasets and can be processed step-by-step accordingly. All data processing complies with the original licenses and ethical guidelines of the source datasets. We are releasing the full data of ID-FFHQ and the processing scripts for ID-CelebA to facilitate future research.
\begin{table}[htbp]
  \centering
  \caption{Basic information of ID-CelebA and ID-FFHQ.}
    \begin{tabular}{l|cc}
    \toprule
    Names  & \multicolumn{1}{c|}{ID-CelebA} & \multicolumn{1}{c}{ID-FFHQ} \\ \hline
Derived From & \multicolumn{1}{c|}{CelebAMaskHQ~\cite{CelebAMask-HQ}} & \multicolumn{1}{c}{FFHQ~\cite{karras2019FFHQ}} \\ \hline
    Image Count & \multicolumn{1}{c|}{18,197} & \multicolumn{1}{c}{25,839} \\ \hline
    Identity Count & \multicolumn{1}{c|}{5,054} & \multicolumn{1}{c}{21,857} \\ \hline
    Resolutions & \multicolumn{1}{c|}{$512\times512$} & \multicolumn{1}{c}{$1024\times1024$} \\ \hline
    Attributes & \multicolumn{2}{c}{Text~\cite{Qwen3-VL}, Sketch~\cite{SimoSerraTOG2018Sketch1, SimoSerraSIGGRAPH2016Sketch2}, $I_{LQ}$, $I_{LQ}'$} \\ \hline
    Saved Form & \multicolumn{2}{c}{.jpg} \\
    \bottomrule
    \end{tabular}%
  \label{tab:dataset}%
\end{table}%

\begin{figure*}[htbp]
\begin{center}
   \includegraphics[width=0.9\linewidth]{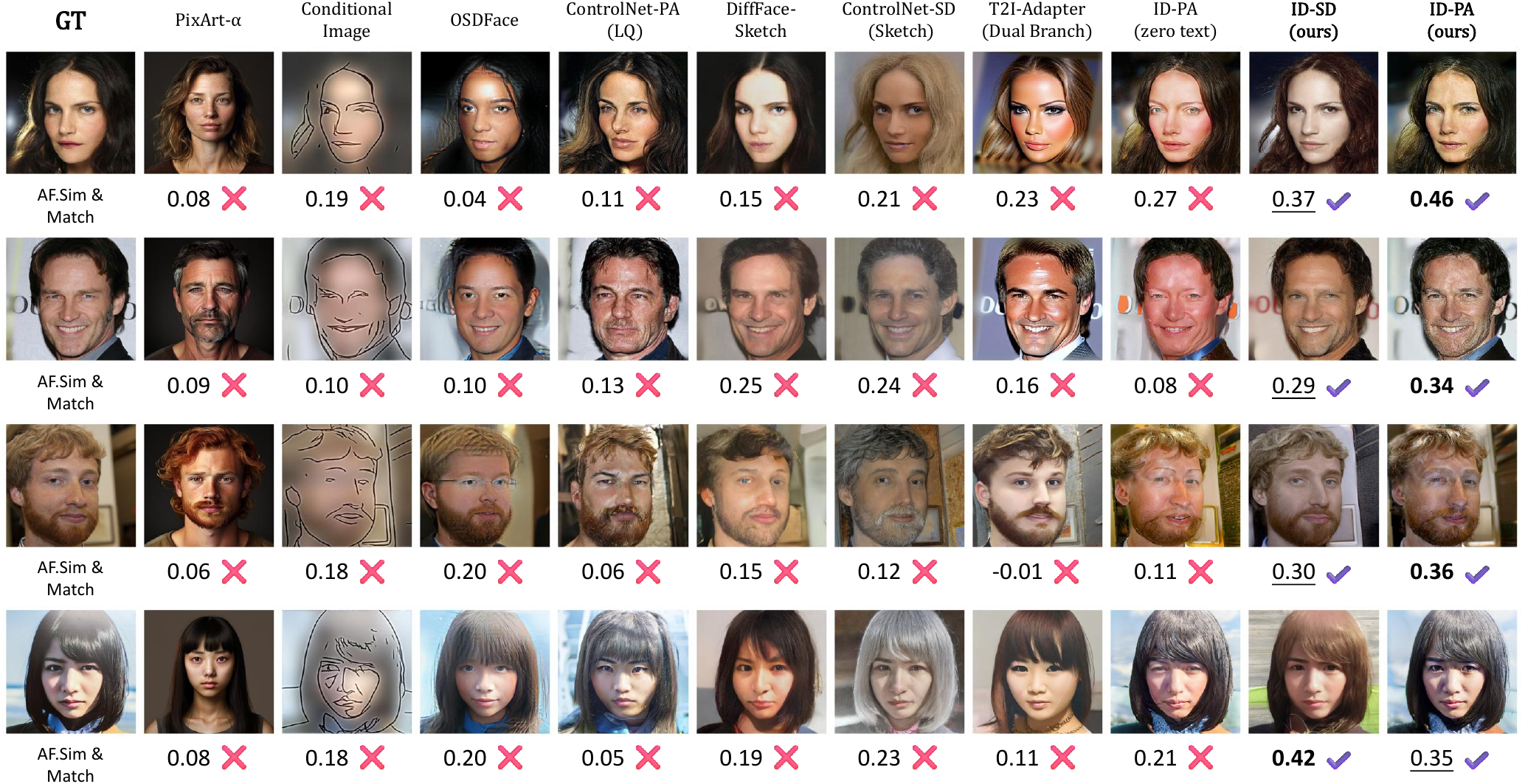}
\end{center}
   \caption{Qualitative comparison of different modality selections and baselines under one-shot generation. IdentiFace achieves the best performances on AdaFace Similarity and matches AdaFace Top-1 Retrieval identity in these examples.} 
\label{fig:quality_modality}
\end{figure*}

\subsection{Image Processing and Identity Construction}
\label{subsec:id_construct}
To simulate real-world suspect scenarios and ensure the generated faces are biometrically identifiable, we applied rigorous filtering and processing protocols.

\textbf{Image Filtering and Accessory Removal.} 
To ensure the quality of facial images, multi-face and non-frontal images were excluded. We employed \textit{Qwen3-VL}~\cite{Qwen3-VL} to filter images based on face count and pose orientation. Additionally, salient accessories, e.g., sunglasses and hat were inpainted to restore the underlying facial structure. 

\textbf{Identity Clustering and Verification.} 
We employed \textit{AdaFace}~\cite{kim2022adaface} as a facial recognition model to construct identity clusters for ID-FFHQ.
Using the existing CelebAMask-HQ identity file, we calculated the cosine similarity threshold of the model $\mathcal{T}_{\text{Ada}} \approx 0.38482$. Applying this threshold to ID-FFHQ resulted in the identity file. 

\subsection{Multi-modal Data Synthesis}
\label{subsec:annotation}
\textbf{Detailed Text Descriptions.} 
Based on forensic facial composite standards, we designed a structured text template covering 8 distinct regions: 
(1) \textit{Basic Info} (gender, perceived age, ethnicity), 
(2) \textit{Overview} (face shape, notable marks), 
(3) \textit{Hair} (length, color, texture), 
(4) \textit{Eyes} (size, shape, iris color), 
(5) \textit{Eyebrows} (thickness, arch), 
(6) \textit{Mouth} (width, lip shape), 
(7) \textit{Nose} (height, width, profile), and 
(8) \textit{Ears} (size, contour). 
The text for each image were generated using \textit{Qwen3.5-VL}~\cite{Qwen3-VL} guided by this template.

\textbf{LQ Image Synthesis.}
LQ images refer to low-quality frontal face images captured by surveillance cameras, with extremely low resolutions. The basic synthesis of $I_{LQ}$ is to resize the GT image to $8\times8$ and back with bicubic interpolation. LQ closer to raw surveillance images, denoted by $I_{LQ}'$, is constructed by adding more camera-like degradations to the original image with a higher resized resolution. In our application scenarios, $I_{LQ}$ is considered accessible through preprocessing $I_{LQ}'$ by technical departments. 

\textbf{Simple Sketch Synthesis.} 
Instead of high-quality ones, we anticipate that sketch is provided by witnesses or artists more simply based on LQ data. We used a pretrained GAN model~\cite{SimoSerraTOG2018Sketch1, SimoSerraSIGGRAPH2016Sketch2} with the processing scripts of an existing work~\cite{xia2021MMCelebA} to generate expected imperfect sketches from GT images. 

\begin{table*}[htbp]
\centering
\caption{Quantitative comparison of modality selections. The two numbers $a/b$ in each cell represent the figures on ID-CelebA and ID-FFHQ respectively, all of which are computed offline after images are saved on disk. We adopted the pretrained models of PixArt-$\alpha$, Stable Diffusion 1.5 and OSDFace, and trained the remaining model instances on our datasets separately. The zero-text version shares the weights of ID-PA and fixes the text input "The photo of a face."}
\begin{subtable}[t]{\textwidth}
\centering
\caption{Comparison on models with pretrained backbone PixArt-$\alpha$~\cite{chen2023pixartalpha} and baselines.}
\footnotesize
\setlength{\tabcolsep}{2.5pt}
\begin{tabular}{l c c c | >{\centering\arraybackslash}>{\centering\arraybackslash}m{4.5em} >{\centering\arraybackslash}>{\centering\arraybackslash}m{4.5em} >{\centering\arraybackslash}>{\centering\arraybackslash}m{4.5em} >{\centering\arraybackslash}>{\centering\arraybackslash}m{4.5em} >{\centering\arraybackslash}>{\centering\arraybackslash}m{4.5em} >{\centering\arraybackslash}>{\centering\arraybackslash}m{4.5em} >{\centering\arraybackslash}>{\centering\arraybackslash}m{4.5em} >{\centering\arraybackslash}>{\centering\arraybackslash}m{4.5em} >{\centering\arraybackslash}>{\centering\arraybackslash}m{4.5em}}
\toprule
\makecell[c]{Models} & \makecell[c]{Text} & \makecell[c]{LQ}& \makecell[c]{Sketch} & 
\makecell[c]{AF. Match\\ Rate$\uparrow$} & 
\makecell[c]{Avg. AF.\\ Sim$\uparrow$} & 
\makecell[c]{SF. Match\\ Rate$\uparrow$} & 
\makecell[c]{Avg. SF.\\ Sim$\uparrow$} & 
\makecell[c]{FID$\downarrow$} & 
\makecell[c]{SR-SIM$\uparrow$} & 
\makecell[c]{LPIPS$\downarrow$} & 
\makecell[c]{MS- \\ SSIM$\uparrow$} & 
\makecell[c]{FG-CLIP\\ Score$\uparrow$} \\ \hline
\makecell[c]{PixArt-$\alpha$ \\ (Model Baseline)} & \makecell[c]{\Checkmark} & \makecell[c]{\XSolidBrush} & \makecell[c]{\XSolidBrush} & 0.00604/ 0.00155 & 0.04378/ 0.03765 & 0.00220/ 0.00077 & 0.18969/ 0.19580 & 84.384/ 90.828 & 0.70460/ 0.69352 & 0.71840/ 0.74031 & 0.37398/ 0.36010 & \textbf{4.5887}/ \textbf{4.9889} \\ \hline
\makecell[c]{Conditional Image \\ (Condition Baseline)} & \makecell[c]{\XSolidBrush} & \makecell[c]{\Checkmark} & \makecell[c]{\Checkmark} & 0.13991/ 0.11056 & 0.13330/ 0.14175 & 0.01648/ 0.00929 & 0.21006/ 0.27753 & 396.18/ 405.33 & 0.76966/ 0.76194 & 0.51342/ 0.53047 & 0.45436/ 0.42766 & N/A \\ \hline
\makecell[c]{ControlNet-PA\\(LQ)} & \makecell[c]{\Checkmark} & \makecell[c]{\Checkmark} & \makecell[c]{\XSolidBrush} & 0.01264/ 0.01278 & 0.08392/ 0.08400 & 0.06374/ 0.05226 & 0.34642/ 0.35450 & 33.526/ 35.947 & 0.78479/ 0.76291 & 0.54789/ 0.57969 & 0.42173/ 0.38681 & 2.4893/ 0.21599 \\ \hline
\makecell[c]{ControlNet-PA\\(Sketch)} & \makecell[c]{\Checkmark} & \makecell[c]{\XSolidBrush} & \makecell[c]{\Checkmark} & \uline{0.33681}/ \uline{0.25513} & \uline{0.22839}/ \uline{0.22261} & \uline{0.48407}/ \uline{0.36663} & \uline{0.49015}/ \uline{0.50490} & 33.399/ 31.860 & 0.83324/ \uline{0.83465} & 0.53158/ 0.57799 & 0.51061/ 0.48905 & \uline{2.6909}/ 0.84734 \\ \hline
\makecell[c]{ID-PA (zero text)} & \makecell[c]{\XSolidBrush} & \makecell[c]{\Checkmark} & \makecell[c]{\Checkmark} & 0.13736/ 0.11774 & 0.16240/ 0.16334 & 0.19670/ 0.23470 & 0.40349/ 0.42226 & 37.819/ 54.828 & \uline{0.83431}/ 0.83261 & 0.51052/ 0.51613 & 0.51879/ 0.50878 & N/A \\ \hline 
\makecell[c]{ControlNet-PA\\(Dual Branch)} & \makecell[c]{\Checkmark} & \makecell[c]{\Checkmark}& \makecell[c]{\Checkmark} & 0.14011/ 0.11541 & 0.17808/ 0.17662 & 0.35659/ 0.28311 & 0.46506/ 0.48659 & \uline{30.698}/ \uline{30.077} & 0.82972/ 0.82641 & \uline{0.48176}/ \uline{0.50177} & \uline{0.52249}/ \uline{0.52901} & 2.6908/ 0.72593 \\ \hline
\makecell[c]{\textbf{ID-PA (ours)}} & \makecell[c]{\Checkmark} & \makecell[c]{\Checkmark} & \makecell[c]{\Checkmark} & \textbf{0.48736}/ \textbf{0.40418} & \textbf{0.27193}/ \textbf{0.26750} & \textbf{0.72198}/ \textbf{0.62060} & \textbf{0.55431}/ \textbf{0.56285} & \textbf{19.775}/ \textbf{27.946} & \textbf{0.86706}/ \textbf{0.85191} & \textbf{0.38949}/ \textbf{0.46314} & \textbf{0.61218}/ \textbf{0.56424} & 2.6368/ \uline{0.97063} \\
\bottomrule
\end{tabular}
\label{subtab:comparison_1}
\end{subtable}
\begin{subtable}[t]{\textwidth}
\centering
\caption{Comparison on models with pretrained backbone SD-1.5, SD-2.1 (OSDFace), and VQF4~\cite{rombach2022high}(DiffFaceSketch).}
\footnotesize
\setlength{\tabcolsep}{2.5pt}
\begin{tabular}{l c c c | >{\centering\arraybackslash}>{\centering\arraybackslash}m{4.5em} >{\centering\arraybackslash}>{\centering\arraybackslash}m{4.5em} >{\centering\arraybackslash}>{\centering\arraybackslash}m{4.5em} >{\centering\arraybackslash}>{\centering\arraybackslash}m{4.5em} >{\centering\arraybackslash}>{\centering\arraybackslash}m{4.5em} >{\centering\arraybackslash}>{\centering\arraybackslash}m{4.5em} >{\centering\arraybackslash}>{\centering\arraybackslash}m{4.5em} >{\centering\arraybackslash}>{\centering\arraybackslash}m{4.5em} >{\centering\arraybackslash}>{\centering\arraybackslash}m{4.5em}}
\toprule
\makecell[c]{Models} & \makecell[c]{Text} & \makecell[c]{LQ}& \makecell[c]{Sketch} & 
\makecell[c]{AF. Match\\ Rate$\uparrow$} & 
\makecell[c]{Avg. AF.\\ Sim$\uparrow$} & 
\makecell[c]{SF. Match\\ Rate$\uparrow$} & 
\makecell[c]{Avg. SF.\\ Sim$\uparrow$} & 
\makecell[c]{FID$\downarrow$} & 
\makecell[c]{SR-SIM$\uparrow$} & 
\makecell[c]{LPIPS$\downarrow$} & 
\makecell[c]{MS- \\ SSIM$\uparrow$} & 
\makecell[c]{FG-CLIP\\ Score$\uparrow$} \\ \hline
\makecell[c]{OSDFace\cite{wang2025osdface}} & \makecell[c]{\XSolidBrush} & \makecell[c]{\Checkmark} & \makecell[c]{\XSolidBrush} & 0.00951/ 0.00894 & 0.05114/ 0.08602 & 0.04341/ 0.04104 & 0.31677/ 0.36127 & 34.237/ 38.645 & 0.79913/ 0.78868 & \uline{0.43225}/ \uline{0.44712} & 0.51352/ 0.50864  & N/A \\ \hline
\makecell[c]{ControlNet-SD\\(LQ)} & \makecell[c]{\Checkmark} & \makecell[c]{\Checkmark} & \makecell[c]{\XSolidBrush} & 0.01319/ 0.00581 & 0.06020/ 0.06440 & 0.04670/ 0.02865 & 0.29301/ 0.30691 & 40.181/ 46.927 & 0.77018/ 0.74155 & 0.56670/ 0.57391 & 0.45455/ 0.41637 & 0.10369/ -1.1448 \\ \hline
\makecell[c]{DiffFaceSketch\cite{peng2023difffacesketch}} & \makecell[c]{\XSolidBrush} & \makecell[c]{\XSolidBrush} & \makecell[c]{\Checkmark} & 0.32727/ 0.23354 & 0.23233/ 0.21234 & 0.52912/ 0.43825 & 0.50755/ 0.50231 & 40.850/ 61.419 & 0.84073/ 0.83270 & \textbf{0.33294}/ \textbf{0.37818} & \uline{0.59556}/ \uline{0.53159} & N/A \\ \hline
\makecell[c]{ControlNet-SD\\(Sketch)} & \makecell[c]{\Checkmark} & \makecell[c]{\XSolidBrush} & \makecell[c]{\Checkmark} & \uline{0.53297}/ \uline{0.44019} & \uline{0.27352}/ \uline{0.27003} & \uline{0.60769}/ \uline{0.51142} & \uline{0.51084}/ \uline{0.52139} & \uline{29.531}/ \uline{32.524} & \uline{0.85269}/ \uline{0.84620} & 0.55084/ 0.52990 & 0.50314/ 0.47581 & 0.25332/ -0.53058 \\ \hline
\makecell[c]{T2I-Adapter\cite{mou2024t2i}\\(Dual Branch)} & \makecell[c]{\Checkmark} & \makecell[c]{\Checkmark} & \makecell[c]{\Checkmark} & 0.06592/ 0.03142 & 0.12623/ 0.11250 & 0.09725/ 0.05923 & 0.38093/ 0.37666 & 62.506/ 36.527 & 0.78089/ 0.79819 & 0.50304/ 0.50900 & 0.44832/ 0.44893 & \textbf{2.2187}/ \textbf{1.2896} \\ \hline 
\makecell[c]{\textbf{ID-SD (ours)}} & \makecell[c]{\Checkmark} & \makecell[c]{\Checkmark} & \makecell[c]{\Checkmark} & \textbf{0.66429}/ \textbf{0.59233} & \textbf{0.30800}/ \textbf{0.31813} & \textbf{0.84780}/ \textbf{0.73558} & \textbf{0.58029}/ \textbf{0.59634} & \textbf{29.142}/ \textbf{29.835} & \textbf{0.86432}/ \textbf{0.86433} & 0.48910/ 0.46939 & \textbf{0.63628}/ \textbf{0.63232} & \uline{0.87088}/ \uline{-0.18050}\\
\bottomrule
\end{tabular}
\label{subtab:comparison_2}
\end{subtable}
\label{tab:comparison}
\end{table*}

\section{Experiments}
\label{sec:experiments}
To make a comprehensive evaluation of the performance of IdentiFace, we conducted experiments on synthetic datasets to validate the methods described in Sec. \ref{sec:methods}, and in real-world scenarios to assess its application capacity in crime investigations.
\subsection{Experimental Setup}
\subsubsection{Basic Setups}
Synthetic-data experiments are conducted on the basis of ID-FFHQ and ID-CelebA. The datasets are divided into training/validation/test sets with a ratio of 80\%/10\%/10\%. To demonstrate the extensibility of IdentiFace, we adopted the official ControlNet~\cite{zhang2023adding} implementations of Stable Diffusion v1.5~\cite{rombach2022high} and PixArt-$\alpha$~\cite{chen2024pixartdelta, chen2023pixartalpha}, which respectively employ a U-Net~\cite{ronneberger2015unet} and DiT~\cite{peebles2023DiT} architecture. We named our IdentiFace instances on them as \textbf{ID-SD} and \textbf{ID-PA}.

We trained the models on multiple NVIDIA RTX Pro 6000 GPUs using the AdamW optimizer. In synthetic-data experiments, we used clean low-res images $I_{LQ}$ as the LQ modality to test the theoretical potential of our method. For facial identity loss $\mathcal{L}_{ID}$, we set $k=10$ and used the pretrained model AdaFace for supervision. Conditional images are resized to $512\times512$ for ID-SD and $1024\times1024$ for ID-PA, and are combined together through a lightweight fusion module with a frequency filter and a residual CNN.

It is worth mentioning that text encoders~\cite{devlin2019bert, radford2021clip, raffel2020t5x} used in diffusion models limit the maximum length of textual input. To facilitate learning of detailed feature descriptions, we divided the structured text for image $I$ into $[Text_{i}^I], i\in\{0,1,\cdots,7\}$, and randomly chose the divided texts as input. Specifically, we define global text, describing overall features, as
\begin{equation}
    {Glo}^I = [Text_{0}^I,Text_{1}^I],
\end{equation}
and local text, focusing on local features, as
\begin{equation}
    Loc^I_j = [Text_{0}^I,Text_{j+2}^I], j \in \{0,1,\cdots,5\}.
\end{equation}
We set a hyperparameter named LTR (Local Text Ratio) in $[0,1]$ that determines the frequency of using local text for training. The default value for LTR is $0.3$, and $j$ is evenly chosen in local texts.

\subsubsection{Evaluation Metrics and Iteration Policy}
\label{subsubsec:user_policy}
We used multiple metrics in three different aspects to evaluate model performances. For identity reconstruction, we provided the average value of Eq. (\ref{eq:similarity}), and identity matching rate of the top-1 retrieval with GT; we used two pretrained models AdaFace~\cite{kim2022adaface} and SFace~\cite{boutros2022sface} to compute them respectively. Specifically, we calculated Average AdaFace \& SFace Similarity (Avg. AF. Sim, Avg. SF. Sim) and AdaFace \& SFace Top-1 Retrieval Match Rate (AF. Match Rate, SF. Match Rate). MTCNN~\cite{zhang2016mtcnn} alignment is applied for AdaFace and is not for SFace in our experiments. Additionally, we included FID~\cite{heusel2017fid}, SR-SIM~\cite{zhang2012sr-sim}, LPIPS~\cite{zhang2018lpips-unreasonable}, MS-SSIM~\cite{wang2003ms-ssim}, and FG-CLIP Score~\cite{xie2025fg-clip} evaluating text-image alignment.

In synthetic-data experiments, we adopted three heuristic rules to simulate the user policy in iterative generation: the simulator (1) generates a face using $Glo^I$, $I_{LQ}$, $I_{Sk}$, then sequentially applies $Loc^I_j$ for editing, with SegFace~\cite{narayan2025segface} providing masks; (2) undoes any edit that lowers AdaFace similarity and fails an AdaFace match; (3) returns from iterations if both AdaFace and SFace ever produce a match during the process. This policy mimics a conservative yet efficient witness behavior.

\subsection{Simulation with Synthetic Data}
\label{subsec:synthetic_experiment}
We conducted comprehensive synthetic-data experiments to validate IdentiFace's core design choices. We evaluated the necessity of three different modalities, the key effect of iterative generation pipeline, and other factors introduced for the task.

\subsubsection{Comparisons on Modality Selections}
In Sec. \ref{subsec:modality}, we proposed three different modalities, LQ, Sketch and text, as a feasible input condition set for identifiable suspect face generation. Results of T2I model~\cite{chen2023pixartalpha} and fused conditional images passing frequency filters are set as baselines. We made comparisons with plain ControlNet models using solely one or two of these conditions, and some specially-designed methods for Sketch-to-Face~\cite{peng2023difffacesketch} and face restoration~\cite{wang2025osdface}. We also made comparisons with T2I-Adapter~\cite{mou2024t2i} and ControlNet, where we used two branches simultaneously to fuse LQ and Sketch conditions. Table \ref{tab:comparison} represents quantitative comparison results of modality selection and combination methods, and Fig. \ref{fig:quality_modality} selected examples of generated images from different methods. All facial images are generated in one shot, without subsequent editing. Compared to other methods, ID-SD and ID-PA show the best performance on identity reconstruction metrics and most other assessment metrics. It indicates that our modality selection and fusion are suitable for this task, and all modalities contribute to generation.

\subsubsection{Iterative Generation Analysis}
\begin{figure}[htbp!]
\begin{center}
   \includegraphics[width=\linewidth]{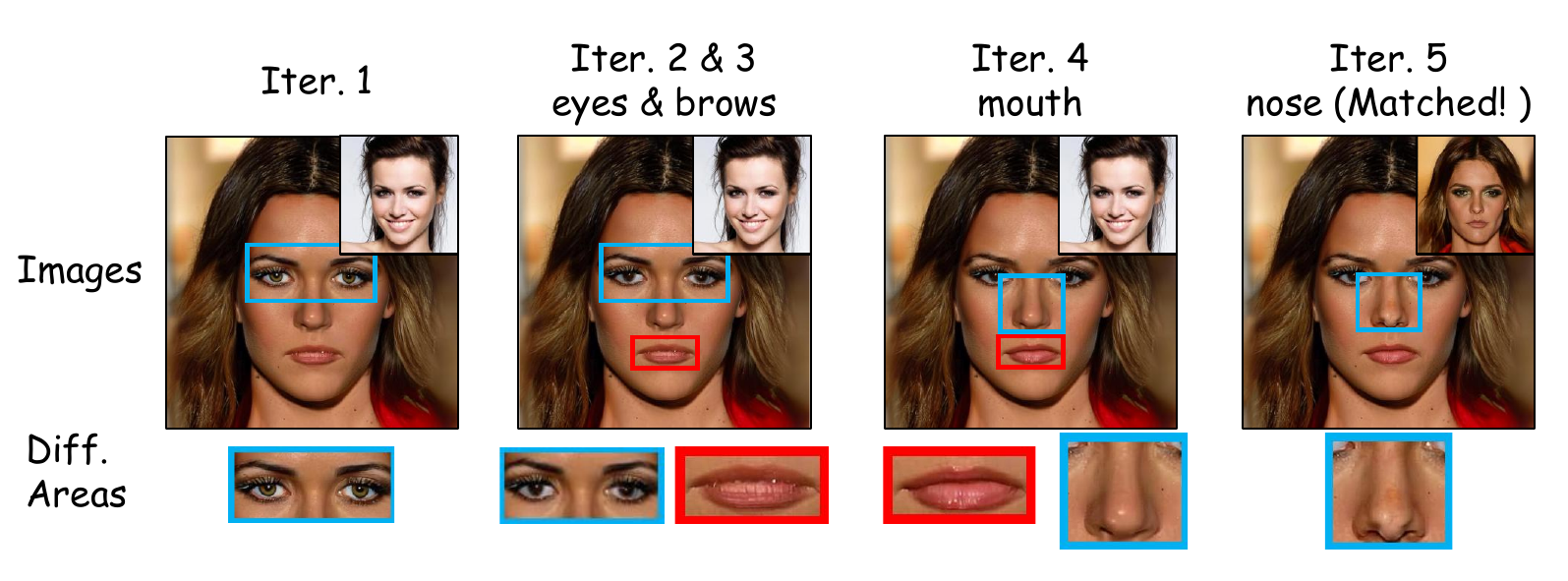}
\end{center}
   \caption{Demonstration of iterative generation process, AdaFace top-1 retrieval images on the top right.} 
\label{fig:iter_demonstration}
\end{figure}

We designed an iterative generation pipeline for better interaction between the user and the model. To evaluate the effect of this pipeline, we conducted large-scale synthetic-data experiments using the heuristic policy raised in Sec. \ref{subsubsec:user_policy}. The iterative generation is tested on both IdentiFace and traditional ControlNet models, and results are shown in Table \ref{tab:comparison_iter} and Fig. \ref{fig:iter_demonstration}. Statistics show that the iteration pipeline significantly improves identity matching rates under appropriate input conditions.

\begin{table}[htbp]
\centering
\caption{Quantitative comparisons between iterative generation (\Checkmark) and one-shot (\XSolidBrush) results. Metrics are calculated online from the generation pipeline, and determine the simulated behaviors. }
\footnotesize
\setlength{\tabcolsep}{2.5pt}
\begin{tabular}{>{\centering\arraybackslash}m{6.5em} l | >{\centering\arraybackslash}m{4.5em} >{\centering\arraybackslash}m{4.5em} >{\centering\arraybackslash}m{4.5em} >{\centering\arraybackslash}m{4.5em}}
\toprule
\makecell[c]{Models} & 
\makecell[c]{Iter. \\ Gen.} & 
\makecell[c]{AF. Match\\ Rate$\uparrow$} & 
\makecell[c]{Avg. AF.\\ Sim$\uparrow$} & 
\makecell[c]{SF. Match\\ Rate$\uparrow$} & 
\makecell[c]{Avg. SF.\\ Sim$\uparrow$} \\ \hline
\multirow{2}{*}{\makecell[c]{ControlNet-SD\\ (Sketch)}} & \makecell[c]{\XSolidBrush} & 0.54178/ 0.51480 & 0.27386/ 0.27082 & 0.49011/ 0.53542 & 0.51900/ 0.54296 \\
                                        & \makecell[c]{\Checkmark} & \textbf{0.75234/ 0.72596}  & \textbf{0.30373/ 0.29962} & \textbf{0.58888/ 0.61628} & \textbf{0.53038/ 0.55378} \\ \hline
\multirow{2}{*}{\makecell[c]{ControlNet-PA\\(Sketch)}} & \makecell[c]{\XSolidBrush} & 0.33480/ 0.27467 & 0.22814/ 0.22254 & 0.48132/ 0.40650 & 0.49026/ 0.50488 \\
                                        & \makecell[c]{\Checkmark} & \textbf{0.55940/ 0.46959} & \textbf{0.26443/ 0.25981} & \textbf{0.56209/ 0.46225} & \textbf{0.50388/ 0.51781} \\ \hline
\multirow{2}{*}{\makecell[c]{ID-SD}} & \makecell[c]{\XSolidBrush} & 0.66482/ 0.69639 & 0.30911/ 0.31838 & 0.65604/ 0.75958 & 0.56981/ 0.60740 \\
                             & \makecell[c]{\Checkmark}  & \textbf{0.83718/ 0.85083} & \textbf{0.33189/ 0.33643}& \textbf{0.72772/ 0.82332} & \textbf{0.57695/ 0.61392}\\ \hline
\multirow{2}{*}{\makecell[c]{ID-PA}} & \makecell[c]{\XSolidBrush} & 0.48653/ 0.44936 & 0.27196/ 0.26796 & 0.71978/ 0.71389 & 0.55438/ 0.56295 \\
                             & \makecell[c]{\Checkmark} & \textbf{0.71978/ 0.66150}& \textbf{0.29861/ 0.29546} & \textbf{0.78187/ 0.76539} & \textbf{0.56377/ 0.57191} \\
\bottomrule
\end{tabular}
\label{tab:comparison_iter}
\end{table}

\subsubsection{Ablations on Other Configurations}
We extensively evaluated how $\mathcal{L}_{ID}$ and LTR influence the model's performance. We trained ID-PA with and without $\mathcal{L}_{ID}$, and for the former we set LTR = 0 and 0.3 respectively. The training and evaluation are based on ID-CelebA dataset. Quantitative results in Table \ref{tab:ablation} imply that both the introduction of $\mathcal{L}_{ID}$ and the diversity of texts enhance the facial feature reconstruction abilities.

\begin{table}[htbp]
\centering
\caption{Ablation studies on LTR values and Facial Identity Loss. }
\footnotesize
\setlength{\tabcolsep}{2.5pt}
\begin{tabular}{>{\centering\arraybackslash}m{3.5em}|>{\centering\arraybackslash}m{2em} l | >{\centering\arraybackslash}m{4.5em} >{\centering\arraybackslash}m{4.5em} >{\centering\arraybackslash}m{4.5em} >{\centering\arraybackslash}m{4.5em}}
\toprule 
\makecell[c]{Iter. \\Gen. }&
\makecell[c]{$\mathcal{L}_{ID}$} & 
\makecell[c]{LTR} &
\makecell[c]{AF. Match\\ Rate$\uparrow$} & 
\makecell[c]{Avg. AF.\\ Sim$\uparrow$} & 
\makecell[c]{SF. Match\\ Rate$\uparrow$} & 
\makecell[c]{Avg. SF.\\ Sim$\uparrow$} \\ \hline
\multirow{3}{*}{\makecell[c]{\XSolidBrush}}& \multirow{2}{*}{\makecell[c]{\Checkmark}}& \makecell[c]{0} & 0.48020 & 0.26704 & 0.70385 & 0.54782 \\
                         
                         & & \makecell[c]{0.3}  & \textbf{0.48653} & \textbf{0.27196} & \textbf{0.71978} & \textbf{0.55438} \\ 
& \makecell[c]{\XSolidBrush} & \makecell[c]{0.3} & 0.44670 & 0.26268 & 0.68736 & 0.54494 \\ \hline
\multirow{3}{*}{\makecell[c]{\Checkmark}} & \multirow{2}{*}{\makecell[c]{\Checkmark}}& \makecell[c]{0} & 0.69214 & 0.29301 & 0.75769 & 0.55385 \\
                        
                         & & \makecell[c]{0.3} & \textbf{0.71978} & \textbf{0.29861} & \textbf{0.78187} & \textbf{0.56377} \\ 
& \makecell[c]{\XSolidBrush} & \makecell[c]{0.3} & 0.68077 & 0.29204 & 0.75110 & 0.55586 \\
\bottomrule
\end{tabular}
\label{tab:ablation}
\end{table}

\subsection{Real-world Experiments}
To evaluate IdentiFace in practical scenarios, we simulated suspect face generation workflows with real volunteers. Seven individuals (S0–S6) provided ID photos as GT to be retrieved, and separate frontal photos as original LQ images. Each LQ image was degraded to a noisy $24\times24$ resolution ($I_{LQ}'$). Collected facial images have been well kept and used only for our experiments. We fine-tuned ID-PA on the ID-FFHQ dataset using $32\times32$ degraded images to learn noise distributions while validating robustness across resolutions.

Eight volunteers participated in generation tasks (S0 for practice, S1–S6 assigned randomly). Instead of using a full sketch $I_{Sk}$, we provided a simpler drawing resembling a sketch synthesized from the witness's memory and perception of $I_{LQ}'$, which needed refinement from volunteers. To simplify the process of facial feature recall by professionals, the GT image was made visible to volunteers as a reference. Volunteers could input text descriptions, modify sketches, provide editing masks, and change random seeds per iteration. Each iteration presented the generated image and the top-1 retrievals of AdaFace and SFace from a facial database containing the 7 samples plus 18,197 faces from ID-CelebA. A trial ended as a success (Pass) when either of the best retrievals matched the target identity, or as a failure (Fail) after 20 minutes.

\begin{table}[htbp]
  \centering
  \caption{Per-sample results of real-world experiments. Total Trials denotes the number of independent generation attempts for S1-S6; Passed Trials indicates how many of them passed; Avg. Iter. is the average number of iterative rounds (including Undo) per trial; Avg. Time is the average time (in minutes) spent per trial. Average values are influenced by testing orders, sample difficulties, strategy preferences, etc., and only passed trials are taken into account.}
  \begin{tabular}{{l}|*{6}{r}}
    \toprule
     Metrics & S1    & S2    & S3    & S4    & S5    & S6 \\ \hline
    \makecell[l]{Total Trials}      &   4     & 4      & 5      & 3      & 4      & 4 \\
    \makecell[l]{Passed Trials}     &   4    & 4     & 5      & 3      & 2      & 2 \\
    \makecell[l]{Avg. Iter.} & 4.00 & 1.25   & 3.20   & 6.33   & 3.00   & 5.50 \\
    \makecell[l]{Avg. Time} & 9.4 & 6.2  & 5.6  & 12.1  & 3.0 & 10.2 \\
    \bottomrule
  \end{tabular}
  \label{tab:real_world}
\end{table}

Table~\ref{tab:real_world} summarizes per‑sample results for S1–S6. The overall pass rate is high, while average time and iteration counts vary in different samples. They are mainly influenced by testing orders, sample difficulties, and volunteers' strategy preferences. For instance, S2 was an easy sample and was often tested early in our experiments, when volunteers were less familiar with the system. As a result, although they succeeded in only one or two iterations, they spent considerable time in the first iteration writing textual descriptions and drawing sketches, which contributed to a longer total trial time per iteration. S5 and S6 are challenging samples. Among the successful trials, volunteers passed S5 in fewer iterations and less time by providing appropriate initial text and sketches. In contrast, those who passed S6 repeatedly refined the input conditions based on the model’s output, revealing a strategic difference in their approaches.

Failed examples occurred in S5 and S6, whose top-1 retrievals continually fell to identical wrong faces respectively, revealing that difficult samples inherently exhibit consistent identity confusion. Nevertheless, the high pass rate and moderate trial time indicate our method's promising potential for practical applications.

\section{Conclusion}
In this work, we proposed IdentiFace, a multi-modal iterative diffusion framework designed for identifiable suspect face generation in crime investigations. We determined multi-modal conditional input, designed an iterative generation pipeline, conducted training loss optimization, and built task-specific datasets. Extensive experiments demonstrated IdentiFace's superior performance over existing methods and real-world application prospects.

Despite the capability of IdentiFace, precise manipulation of fine‑grained attributes might yield unnatural results or require multiple attempts, and bias in training data could affect performance on certain ethnic groups. It also raises potential privacy and misuse concerns. Future work can focus on localized controllability and training-data fairness.

{\small
\bibliographystyle{ieee}
\bibliography{bibs}

@article{ho2020denoising,
  title={Denoising diffusion probabilistic models},
  author={Ho, Jonathan and Jain, Ajay and Abbeel, Pieter},
  journal={Advances in neural information processing systems},
  volume={33},
  pages={6840--6851},
  year={2020}
}

@inproceedings{rombach2022high,
  title={High-resolution image synthesis with latent diffusion models},
  author={Rombach, Robin and Blattmann, Andreas and Lorenz, Dominik and Esser, Patrick and Ommer, Bj{\"o}rn},
  booktitle={Proceedings of the IEEE/CVF conference on computer vision and pattern recognition},
  pages={10684--10695},
  year={2022}
}

@misc{chen2023pixartalpha,
      title={PixArt-$\alpha$: Fast Training of Diffusion Transformer for Photorealistic Text-to-Image Synthesis}, 
      author={Junsong Chen and Jincheng Yu and Chongjian Ge and Lewei Yao and Enze Xie and Yue Wu and Zhongdao Wang and James Kwok and Ping Luo and Huchuan Lu and Zhenguo Li},
      year={2023},
      eprint={2310.00426},
      archivePrefix={arXiv},
      primaryClass={cs.CV}
}

@misc{chen2024pixartdelta,
      title={PIXART-{$\delta$}: Fast and Controllable Image Generation with Latent Consistency Models}, 
      author={Junsong Chen and Yue Wu and Simian Luo and Enze Xie and Sayak Paul and Ping Luo and Hang Zhao and Zhenguo Li},
      year={2024},
      eprint={2401.05252},
      archivePrefix={arXiv},
      primaryClass={cs.CV}
}

@article{cai2025z,
  title={Z-image: An efficient image generation foundation model with single-stream diffusion transformer},
  author={Cai, Huanqia and Cao, Sihan and Du, Ruoyi and Gao, Peng and Hoi, Steven and Hou, Zhaohui and Huang, Shijie and Jiang, Dengyang and Jin, Xin and Li, Liangchen and others},
  journal={arXiv preprint arXiv:2511.22699},
  year={2025}
}

@misc{xie2025sana,
      title={SANA 1.5: Efficient Scaling of Training-Time and Inference-Time Compute in Linear Diffusion Transformer},
      author={Xie, Enze and Chen, Junsong and Zhao, Yuyang and Yu, Jincheng and Zhu, Ligeng and Lin, Yujun and Zhang, Zhekai and Li, Muyang and Chen, Junyu and Cai, Han and others},
      year={2025},
      eprint={2501.18427},
      archivePrefix={arXiv},
      primaryClass={cs.CV},
      url={https://arxiv.org/abs/2501.18427},
    }

@inproceedings{zhang2023adding,
  title={Adding conditional control to text-to-image diffusion models},
  author={Zhang, Lvmin and Rao, Anyi and Agrawala, Maneesh},
  booktitle={Proceedings of the IEEE/CVF international conference on computer vision},
  pages={3836--3847},
  year={2023}
}

@inproceedings{wang2025osdface,
  title={Osdface: One-step diffusion model for face restoration},
  author={Wang, Jingkai and Gong, Jue and Zhang, Lin and Chen, Zheng and Liu, Xing and Gu, Hong and Liu, Yutong and Zhang, Yulun and Yang, Xiaokang},
  booktitle={Proceedings of the Computer Vision and Pattern Recognition Conference},
  pages={12626--12636},
  year={2025}
}

@article{peng2023difffacesketch,
  title={Difffacesketch: High-fidelity face image synthesis with sketch-guided latent diffusion model},
  author={Peng, Yichen and Zhao, Chunqi and Xie, Haoran and Fukusato, Tsukasa and Miyata, Kazunori},
  journal={arXiv preprint arXiv:2302.06908},
  year={2023}
}

@inproceedings{peebles2023DiT,
  title={Scalable diffusion models with transformers},
  author={Peebles, William and Xie, Saining},
  booktitle={Proceedings of the IEEE/CVF international conference on computer vision},
  pages={4195--4205},
  year={2023}
}

@inproceedings{ronneberger2015unet,
  title={U-net: Convolutional networks for biomedical image segmentation},
  author={Ronneberger, Olaf and Fischer, Philipp and Brox, Thomas},
  booktitle={International Conference on Medical image computing and computer-assisted intervention},
  pages={234--241},
  year={2015},
  organization={Springer}
}

@inproceedings{mou2024t2i,
  title={T2i-adapter: Learning adapters to dig out more controllable ability for text-to-image diffusion models},
  author={Mou, Chong and Wang, Xintao and Xie, Liangbin and Wu, Yanze and Zhang, Jian and Qi, Zhongang and Shan, Ying},
  booktitle={Proceedings of the AAAI conference on artificial intelligence},
  volume={38},
  number={5},
  pages={4296--4304},
  year={2024}
}

@inproceedings{duan2025dit4sr,
  title={Dit4sr: Taming diffusion transformer for real-world image super-resolution},
  author={Duan, Zheng-Peng and Zhang, Jiawei and Jin, Xin and Zhang, Ziheng and Xiong, Zheng and Zou, Dongqing and Ren, Jimmy S and Guo, Chunle and Li, Chongyi},
  booktitle={Proceedings of the IEEE/CVF International Conference on Computer Vision},
  pages={18948--18958},
  year={2025}
}

@misc{labs2025flux1kontextflowmatching,
      title={FLUX.1 Kontext: Flow Matching for In-Context Image Generation and Editing in Latent Space},
      author={Black Forest Labs and Stephen Batifol and Andreas Blattmann and Frederic Boesel and Saksham Consul and Cyril Diagne and Tim Dockhorn and Jack English and Zion English and Patrick Esser and Sumith Kulal and Kyle Lacey and Yam Levi and Cheng Li and Dominik Lorenz and Jonas Müller and Dustin Podell and Robin Rombach and Harry Saini and Axel Sauer and Luke Smith},
      year={2025},
      eprint={2506.15742},
      archivePrefix={arXiv},
      primaryClass={cs.GR},
      url={https://arxiv.org/abs/2506.15742},
}

@inproceedings{radford2021clip,
  title={Learning transferable visual models from natural language supervision},
  author={Radford, Alec and Kim, Jong Wook and Hallacy, Chris and Ramesh, Aditya and Goh, Gabriel and Agarwal, Sandhini and Sastry, Girish and Askell, Amanda and Mishkin, Pamela and Clark, Jack and others},
  booktitle={International conference on machine learning},
  pages={8748--8763},
  year={2021},
  organization={PmLR}
}

@article{tang2024toward_SKETCH2FACE,
  title={Toward identity preserving in face sketch-photo synthesis using a hybrid CNN-Mamba framework},
  author={Tang, Duoxun and Jiang, Xinhang and Wang, Kunpeng and Guo, Weichen and Zhang, Jingyuan and Lin, Ye and Pu, Haibo},
  journal={Scientific Reports},
  volume={14},
  number={1},
  pages={22495},
  year={2024},
  publisher={Nature Publishing Group UK London}
}

@article{couairon2022diffedit,
  title={Diffedit: Diffusion-based semantic image editing with mask guidance},
  author={Couairon, Guillaume and Verbeek, Jakob and Schwenk, Holger and Cord, Matthieu},
  journal={arXiv preprint arXiv:2210.11427},
  year={2022}
}

@inproceedings{kim2022adaface,
  title={Adaface: Quality adaptive margin for face recognition},
  author={Kim, Minchul and Jain, Anil K and Liu, Xiaoming},
  booktitle={Proceedings of the IEEE/CVF conference on computer vision and pattern recognition},
  pages={18750--18759},
  year={2022}
}

@inproceedings{boutros2022sface,
  title={Sface: Privacy-friendly and accurate face recognition using synthetic data},
  author={Boutros, Fadi and Huber, Marco and Siebke, Patrick and Rieber, Tim and Damer, Naser},
  booktitle={2022 IEEE International Joint Conference on Biometrics (IJCB)},
  pages={1--11},
  year={2022},
  organization={IEEE}
}

@inproceedings{CelebAMask-HQ,
  title={MaskGAN: Towards Diverse and Interactive Facial Image Manipulation},
  author={Lee, Cheng-Han and Liu, Ziwei and Wu, Lingyun and Luo, Ping},
  booktitle={IEEE Conference on Computer Vision and Pattern Recognition (CVPR)},
  year={2020}
}

@inproceedings{liu2015CelebA,
  title={Deep learning face attributes in the wild},
  author={Liu, Ziwei and Luo, Ping and Wang, Xiaogang and Tang, Xiaoou},
  booktitle={Proceedings of the IEEE international conference on computer vision},
  pages={3730--3738},
  year={2015}
}

@article{karras2017CelebAHQ,
  title={Progressive growing of gans for improved quality, stability, and variation},
  author={Karras, Tero and Aila, Timo and Laine, Samuli and Lehtinen, Jaakko},
  journal={arXiv preprint arXiv:1710.10196},
  year={2017}
}

@inproceedings{karras2019FFHQ,
  title={A style-based generator architecture for generative adversarial networks},
  author={Karras, Tero and Laine, Samuli and Aila, Timo},
  booktitle={Proceedings of the IEEE/CVF conference on computer vision and pattern recognition},
  pages={4401--4410},
  year={2019}
}

@inproceedings{xia2021MMCelebA,
  title={Tedigan: Text-guided diverse face image generation and manipulation},
  author={Xia, Weihao and Yang, Yujiu and Xue, Jing-Hao and Wu, Baoyuan},
  booktitle={Proceedings of the IEEE/CVF conference on computer vision and pattern recognition},
  pages={2256--2265},
  year={2021}
}

@Article{SimoSerraTOG2018Sketch1,
   author    = {Edgar Simo-Serra and Satoshi Iizuka and Hiroshi Ishikawa},
   title     = {{Mastering Sketching: Adversarial Augmentation for Structured Prediction}},
   journal   = "ACM Transactions on Graphics (TOG)",
   year      = 2018,
   volume    = 37,
   number    = 1,
}

@Article{SimoSerraSIGGRAPH2016Sketch2,
   author    = {Edgar Simo-Serra and Satoshi Iizuka and Kazuma Sasaki and Hiroshi Ishikawa},
   title     = {{Learning to Simplify: Fully Convolutional Networks for Rough Sketch Cleanup}},
   journal   = "ACM Transactions on Graphics (SIGGRAPH)",
   year      = 2016,
   volume    = 35,
   number    = 4,
}

@article{Qwen3-VL,
      title={Qwen3-VL Technical Report}, 
      author={Shuai Bai and Yuxuan Cai and Ruizhe Chen and Keqin Chen and Xionghui Chen and Zesen Cheng and Lianghao Deng and Wei Ding and Chang Gao and Chunjiang Ge and Wenbin Ge and Zhifang Guo and Qidong Huang and Jie Huang and Fei Huang and Binyuan Hui and Shutong Jiang and Zhaohai Li and Mingsheng Li and Mei Li and Kaixin Li and Zicheng Lin and Junyang Lin and Xuejing Liu and Jiawei Liu and Chenglong Liu and Yang Liu and Dayiheng Liu and Shixuan Liu and Dunjie Lu and Ruilin Luo and Chenxu Lv and Rui Men and Lingchen Meng and Xuancheng Ren and Xingzhang Ren and Sibo Song and Yuchong Sun and Jun Tang and Jianhong Tu and Jianqiang Wan and Peng Wang and Pengfei Wang and Qiuyue Wang and Yuxuan Wang and Tianbao Xie and Yiheng Xu and Haiyang Xu and Jin Xu and Zhibo Yang and Mingkun Yang and Jianxin Yang and An Yang and Bowen Yu and Fei Zhang and Hang Zhang and Xi Zhang and Bo Zheng and Humen Zhong and Jingren Zhou and Fan Zhou and Jing Zhou and Yuanzhi Zhu and Ke Zhu},
	  journal={arXiv preprint arXiv:2511.21631},
      year={2025}
}

@article{heusel2017fid,
  title={Gans trained by a two time-scale update rule converge to a local nash equilibrium},
  author={Heusel, Martin and Ramsauer, Hubert and Unterthiner, Thomas and Nessler, Bernhard and Hochreiter, Sepp},
  journal={Advances in neural information processing systems},
  volume={30},
  year={2017}
}

@inproceedings{wang2003ms-ssim,
  title={Multiscale structural similarity for image quality assessment},
  author={Wang, Zhou and Simoncelli, Eero P and Bovik, Alan C},
  booktitle={The thrity-seventh asilomar conference on signals, systems \& computers, 2003},
  volume={2},
  pages={1398--1402},
  year={2003},
  organization={Ieee}
}

@inproceedings{zhang2012sr-sim,
  title={SR-SIM: A fast and high performance IQA index based on spectral residual},
  author={Zhang, Lin and Li, Hongyu},
  booktitle={2012 19th IEEE international conference on image processing},
  pages={1473--1476},
  year={2012},
  organization={IEEE}
}

@inproceedings{zhang2018lpips-unreasonable,
  title={The unreasonable effectiveness of deep features as a perceptual metric},
  author={Zhang, Richard and Isola, Phillip and Efros, Alexei A and Shechtman, Eli and Wang, Oliver},
  booktitle={Proceedings of the IEEE conference on computer vision and pattern recognition},
  pages={586--595},
  year={2018}
}

@article{xie2025fg-clip,
  title={Fg-clip: Fine-grained visual and textual alignment},
  author={Xie, Chunyu and Wang, Bin and Kong, Fanjing and Li, Jincheng and Liang, Dawei and Zhang, Gengshen and Leng, Dawei and Yin, Yuhui},
  journal={arXiv preprint arXiv:2505.05071},
  year={2025}
}

@inproceedings{narayan2025segface,
  title={Segface: Face segmentation of long-tail classes},
  author={Narayan, Kartik and Vs, Vibashan and Patel, Vishal M},
  booktitle={Proceedings of the AAAI Conference on Artificial Intelligence},
  volume={39},
  number={6},
  pages={6182--6190},
  year={2025}
}

@inproceedings{deng2019arcface,
  title={Arcface: Additive angular margin loss for deep face recognition},
  author={Deng, Jiankang and Guo, Jia and Xue, Niannan and Zafeiriou, Stefanos},
  booktitle={Proceedings of the IEEE/CVF conference on computer vision and pattern recognition},
  pages={4690--4699},
  year={2019}
}

@inproceedings{warrier2024criminal-stable,
  title={Generation and editing of faces using stable diffusion with criminal suspect matching},
  author={Warrier, Aditya and Mathew, Amisha and Patra, Amrita and Hiremath, Khushi S and Jijo, Jeny},
  booktitle={2024 IEEE International Conference on Advanced Systems and Emergent Technologies (IC\_ASET)},
  pages={1--6},
  year={2024},
  organization={IEEE}
}

@inproceedings{kulkarni2025criminal-gan,
  title={Applying GANs for Image Synthesis and Recognition in Forensic Contexts},
  author={Kulkarni, Vinaya and Karande, Shital and Patil, Jagruti and Adhikari, Anushruti and Jariwala, Diti and Nigade, Arya},
  booktitle={2025 12th International Conference on Computing for Sustainable Global Development (INDIACom)},
  pages={1--6},
  year={2025},
  organization={IEEE}
}

@article{saharia2022imagen,
  title={Photorealistic text-to-image diffusion models with deep language understanding},
  author={Saharia, Chitwan and Chan, William and Saxena, Saurabh and Li, Lala and Whang, Jay and Denton, Emily L and Ghasemipour, Kamyar and Gontijo Lopes, Raphael and Karagol Ayan, Burcu and Salimans, Tim and others},
  journal={Advances in neural information processing systems},
  volume={35},
  pages={36479--36494},
  year={2022}
}

@article{raffel2020t5x,
  title={Exploring the limits of transfer learning with a unified text-to-text transformer},
  author={Raffel, Colin and Shazeer, Noam and Roberts, Adam and Lee, Katherine and Narang, Sharan and Matena, Michael and Zhou, Yanqi and Li, Wei and Liu, Peter J},
  journal={Journal of machine learning research},
  volume={21},
  number={140},
  pages={1--67},
  year={2020}
}

@inproceedings{devlin2019bert,
  title={Bert: Pre-training of deep bidirectional transformers for language understanding},
  author={Devlin, Jacob and Chang, Ming-Wei and Lee, Kenton and Toutanova, Kristina},
  booktitle={Proceedings of the 2019 conference of the North American chapter of the association for computational linguistics: human language technologies, volume 1 (long and short papers)},
  pages={4171--4186},
  year={2019}
}

@article{zhang2016mtcnn,
  title={Joint face detection and alignment using multitask cascaded convolutional networks},
  author={Zhang, Kaipeng and Zhang, Zhanpeng and Li, Zhifeng and Qiao, Yu},
  journal={IEEE signal processing letters},
  volume={23},
  number={10},
  pages={1499--1503},
  year={2016},
  publisher={IEEE}
}

@article{song2020denoising,
  title={Denoising diffusion implicit models},
  author={Song, Jiaming and Meng, Chenlin and Ermon, Stefano},
  journal={arXiv preprint arXiv:2010.02502},
  year={2020}
}

@article{qin2023unicontrol,
  title={Unicontrol: A unified diffusion model for controllable visual generation in the wild},
  author={Qin, Can and Zhang, Shu and Yu, Ning and Feng, Yihao and Yang, Xinyi and Zhou, Yingbo and Wang, Huan and Niebles, Juan Carlos and Xiong, Caiming and Savarese, Silvio and others},
  journal={arXiv preprint arXiv:2305.11147},
  year={2023}
}

@article{zhao2023uni,
  title={Uni-controlnet: All-in-one control to text-to-image diffusion models},
  author={Zhao, Shihao and Chen, Dongdong and Chen, Yen-Chun and Bao, Jianmin and Hao, Shaozhe and Yuan, Lu and Wong, Kwan-Yee K},
  journal={Advances in neural information processing systems},
  volume={36},
  pages={11127--11150},
  year={2023}
}

@article{avrahami2023blended,
  title={Blended latent diffusion},
  author={Avrahami, Omri and Fried, Ohad and Lischinski, Dani},
  journal={ACM transactions on graphics (TOG)},
  volume={42},
  number={4},
  pages={1--11},
  year={2023},
  publisher={ACM New York, NY, USA}
}

@inproceedings{avrahami2022blended,
  title={Blended diffusion for text-driven editing of natural images},
  author={Avrahami, Omri and Lischinski, Dani and Fried, Ohad},
  booktitle={Proceedings of the IEEE/CVF conference on computer vision and pattern recognition},
  pages={18208--18218},
  year={2022}
}

@inproceedings{wang2023imagenEditor,
  title={Imagen editor and editbench: Advancing and evaluating text-guided image inpainting},
  author={Wang, Su and Saharia, Chitwan and Montgomery, Ceslee and Pont-Tuset, Jordi and Noy, Shai and Pellegrini, Stefano and Onoe, Yasumasa and Laszlo, Sarah and Fleet, David J and Soricut, Radu and others},
  booktitle={Proceedings of the IEEE/CVF conference on computer vision and pattern recognition},
  pages={18359--18369},
  year={2023}
}

@article{wang2008cuhkDataset,
  title={Face photo-sketch synthesis and recognition},
  author={Wang, Xiaogang and Tang, Xiaoou},
  journal={IEEE transactions on pattern analysis and machine intelligence},
  volume={31},
  number={11},
  pages={1955--1967},
  year={2008},
  publisher={IEEE}
}

@inproceedings{ravi2024face,
  title={Face Generation and Recognition in Forensic Science},
  author={Ravi, Gayathri and Joy, Heynes and Jitto, Jeffin and Joshy, Jocelyn and Jose, Jisha Mary},
  booktitle={2024 11th International Conference on Advances in Computing and Communications (ICACC)},
  pages={1--4},
  year={2024},
  organization={IEEE}
}

@inproceedings{jalan2020suspect,
  title={Suspect face generation},
  author={Jalan, Harsh Jaykumar and Maurya, Gautam and Corda, Canute and Dsouza, Sunny and Panchal, Dakshata},
  booktitle={2020 3rd International Conference on Communication System, Computing and IT Applications (CSCITA)},
  pages={73--78},
  year={2020},
  organization={IEEE}
}

@article{que2024denoising,
  title={Denoising diffusion probabilistic model for face sketch-to-photo synthesis},
  author={Que, Yue and Xiong, Li and Wan, Weiguo and Xia, Xue and Liu, Zhiwei},
  journal={IEEE Transactions on Circuits and Systems for Video Technology},
  volume={34},
  number={10},
  pages={10424--10436},
  year={2024},
  publisher={IEEE}
}
}

\end{document}